\documentclass[conference]{IEEEtran}
\IEEEoverridecommandlockouts
\usepackage{cite}
\usepackage{amsmath,amssymb,amsfonts}
\usepackage{graphicx}
\usepackage{textcomp}
\usepackage{xcolor}
\def\BibTeX{{\rm B\kern-.05em{\sc i\kern-.025em b}\kern-.08em
    T\kern-.1667em\lower.7ex\hbox{E}\kern-.125emX}}

\usepackage{amsmath}
\usepackage{algorithm}
\usepackage{siunitx}
\usepackage[noend]{algpseudocode}
\usepackage{multicol}
\usepackage{wrapfig}
\usepackage{subfig}
\usepackage{mathrsfs}
\usepackage[bookmarks=false]{hyperref}
\usepackage{color}
\usepackage{listings}
\usepackage{dblfloatfix}
\usepackage{tikz}

\newcommand\copyrighttext{%
  \footnotesize © 2020 IEEE.  Personal use of this material is permitted.  Permission from IEEE must be obtained for all other uses, in any current or future media, including reprinting/republishing this material for advertising or promotional purposes, creating new collective works, for resale or redistribution to servers or lists, or reuse of any copyrighted component of this work in other works.}
\newcommand\copyrightnotice{%
\begin{tikzpicture}[remember picture,overlay]
\node[anchor=south,yshift=10pt] at (current page.south) {\fbox{\parbox{\dimexpr\textwidth-\fboxsep-\fboxrule\relax}{\copyrighttext}}};
\end{tikzpicture}%
}

\def\BibTeX{{\rm B\kern-.05em{\sc i\kern-.025em b}\kern-.08em
    T\kern-.1667em\lower.7ex\hbox{E}\kern-.125emX}}
\begin{document}

\title{Building an Automated and Self-Aware Anomaly Detection System}

\author{\IEEEauthorblockN{Sayan Chakraborty}
\IEEEauthorblockA{\textit{Zillow Group} \\
Seattle, USA \\
sayanc@zillowgroup.com}\\

\IEEEauthorblockN{Anna Swigart}
\IEEEauthorblockA{\textit{Zillow Group} \\
Seattle, USA \\
annasw@zillowgroup.com}

\and
\IEEEauthorblockN{Smit Shah}
\IEEEauthorblockA{\textit{Zillow Group} \\
Seattle, USA \\
smits@zillowgroup.com}\\

\IEEEauthorblockN{Luyao Yang}
\IEEEauthorblockA{\textit{Zillow Group} \\
Seattle, USA \\
luyaoy@zillowgroup.com}

\and
\IEEEauthorblockN{Kiumars Soltani}
\IEEEauthorblockA{\textit{Zillow Group} \\
Seattle, USA \\
kiumarss@zillowgroup.com}\\

\IEEEauthorblockN{Kyle Buckingham}
\IEEEauthorblockA{\textit{Zillow Group} \\
Seattle, USA \\
kylebu@zillowgroup.com}
}

\maketitle

\begin{abstract}
Organizations rely heavily on time series metrics to measure and model key aspects of operational and business performance. The ability to reliably detect issues with these metrics is imperative to identifying early indicators of major problems before they become pervasive. It can be very challenging to proactively monitor a large number of diverse and constantly changing time series for anomalies, so there are often gaps in monitoring coverage, disabled or ignored monitors due to false positive alarms, and teams resorting to manual inspection of charts to catch problems. Traditionally, variations in the data generation processes and patterns have required strong modeling expertise to create models that accurately flag anomalies. In this paper, we describe an anomaly detection system that overcomes this common challenge by keeping track of its own performance and making changes as necessary to each model without requiring manual intervention. We demonstrate that this novel approach outperforms available alternatives on benchmark datasets in many scenarios.

\end{abstract}

\copyrightnotice


\section{Introduction}
Detecting data quality issues is a challenging yet crucial operation within every data-driven organization. Companies are dedicating lots of resources to identify data quality issues in a time-sensitive fashion to avoid cascading failure effects. Previous work has widely used anomaly detection methods to monitor time series data. However, there are challenges remaining with how to scale such methods across different teams and business domains.

To effectively monitor data within a large ecosystem, an accurate and efficient approach that can be easily adapted for different use cases is required. This is particularly challenging when metrics are generated by different internal or external processes, with varying seasonalities, granularity, and noise. With these conditions, using manual model development and training is expensive and ineffective.

In addition to scaling the data quality monitoring coverage within an organization, we need an approach that can be deployed with minimal configuration, allowing the process of modeling, training and configuration tuning to be automated.

\begin{figure}[H]
\begin{center}
\includegraphics[ width=0.48\textwidth]{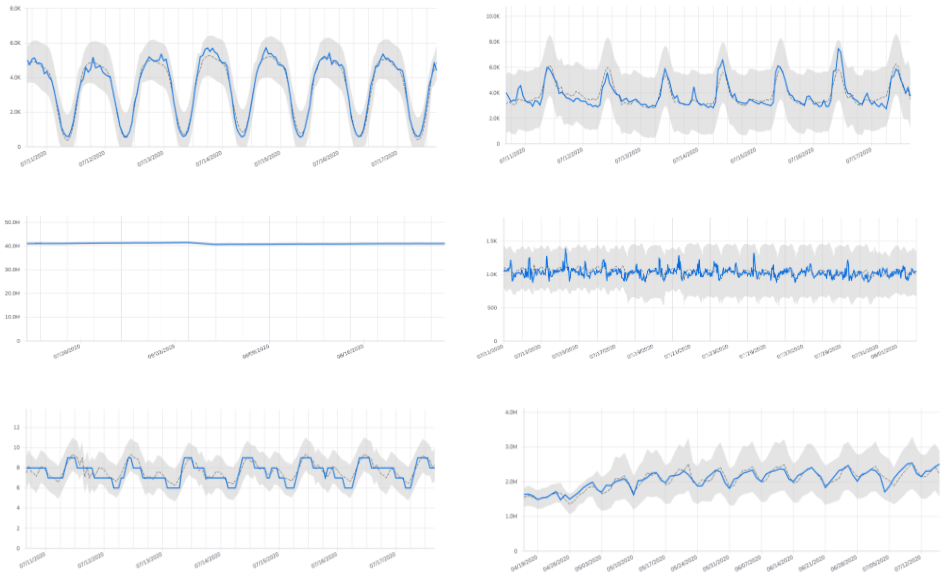}
\caption{Time series with variant structural patterns}
\label{various_charts}
\end{center}
\end{figure}

In order to implement an effective anomaly detection approach, we have identified the following challenges that need to be addressed:
\begin{itemize}
\item \textbf{Generalization:} Time series data patterns and characteristics are largely variant (see figure \ref{various_charts}), depending on the data generation processes and business domain. Therefore, the anomaly detection approach should be \textit{generalizable} to incorporate a wide range of business and operational metrics (\cite{ren2019time}).
\item \textbf{Automation:} Configuration of the right anomaly detection model varies a lot from one problem to another, requiring strong domain expertise and manual tuning which is both time and labor consuming. Therefore, exhaustive monitoring coverage requires automating the process of data collection, modeling and scoring as much as possible to increase adoption of the solution within an organization.
\item \textbf{Self-awareness:} Variational patterns of the data generated by a system may evolve over time. Hence, the anomaly detector needs to seamlessly adapt to such changes without compromising its anomaly classification performance.
\item \textbf{Democratization:} Visibility of monitoring across multiple system components is important to identify the root cause of a give issue. Independent monitoring over different components creates complexity in terms of visibility and interpretations. An effective approach should \textit{democratize} the anomaly detection and reporting mechanism by providing a self-service onboarding tool that only requires minimal configuration and hides the detail of the ML backend from the users.
\end{itemize}

To address the aforementioned challenges, we have designed and implemented an automated anomaly detection system across Zillow Group that enables scientists, engineers and product managers to easily create machine learning powered pipelines to monitor their metrics and receive alerts when anomalies are observed.

Our system is backed by a hybrid unsupervised algorithm that includes two main classes of anomaly detection models (structure based and filter-based) and automatically select and adjust the best model for the user's data. We achieve this by modeling anomaly detection as an optimization problem with a multi term objective function that measures both predictive accuracy and uncertainty fluctuations.

In addition, we have employed a self-awareness component to address changes in the underlying data, such as structural or variational changes, as well as model misspecification during the initial training phase. Our approach measures evaluation metrics for unsupervised models such as Mass Volume (MV) curve and Excess Mass (EM) curve with some other naive metrics such as the rate of anomalies per scoring instance as the performance indicators of our anomaly detection approach and trigger the auto-tuning process if a significant change is observed.

To democratize the data monitoring process, we have deployed our system as a self-service platform (refer to Section \ref{sec:system}) that allows users to interact with a wrapper around the configuration and management of the core ML service. By taking advantage of automated configuration tuning, we are able to hide the complexity and details of our ML backend from the user. This has played a crucial role in adoption of our platform within different teams in our company.

We evaluate our approach by comparing its anomaly classification and forecasting performance with three well known time series anomaly detection/forecasting solutions: Prophet, Luminol and ADTK (Section \ref{sec:benchmarks}). To run the experiments, we use benchmark data from Numenta Anomaly Benchmark (NAB). Our experiments show that our approach can reach the same level of performance, and often outperforms these existing solutions, without any human intervention.

\section{Related Works} \label{sec:relatedwork}
Organizations must incorporate relevant monitoring at each phase of a data pipeline in order to ensure its stability. Typically, a specific process output is an aggregated result of a series of upstream processes. Hence, the root of a cascading failure can be caught if a detailed mapping of all the dependent process call graphs are available (\cite{kim2013root}). Moreover, some accumulated minor upstream changes that go undetected may lead to serious issues over time for a downstream process (e.g. a shift in feature distribution that might invalidate a configuration of a downstream machine learning model) (\cite{sculley2015hidden}, \cite{diethe2019continual}). One solution to detect such problems is to correlate the anomaly scores at several component levels (\cite{Marwede2009auto}). However, the success of such an approach depends on an exhaustive coverage of monitoring through different system components.

When developing an anomaly detection model, it is important to be able to capture what constitutes an unexpected value for a specific metric or dataset. Gathering profile information for batches of data can help summarize any major data shifts that should be accounted for in training. A typical approach for this is to compute information related to past change points or trend changes (\cite{liu2013change}, \cite{lund2007changepoint}, \cite{kawahara2007change}). For online anomaly detection use cases, the types of anomalies vary a lot from one use case to another, but typically can be defined as the point or a set of points that behave significantly different from the others. There are several methods available that can track outlying points for low frequency time series data (\cite{soule2005combining}, \cite{malhotra2015long}, \cite{barz2018detecting}, \cite{liu2004line}) and for streaming use cases (\cite{hill2010anomaly}, \cite{lakhina2004characterization}, \cite{ahmad2017unsupervised}). These methods have very minimal requirements for external information outside of the time series data, and depend mostly on quantifying the patterns present within the time series itself. However, methods exist to allow the flexibility of adding external information in the model (such as holidays) (\cite{zhu2017deep}, \cite{gupta2013context}) for improved predictive performance and for preventing false positives when fluctuations are expected. Moreover, there are techniques that go a step further to find causal relationships between time series to identify the actual root cause for the an unexpected event (\cite{chakraborty2019root}, \cite{qiu2012granger}, \cite{silveira2010urca}, \cite{liu2016root}).

As mentioned in the previous section, complexities of different services and their interactions at an organizational level require an off-the-shelf system that can perform anomaly detection for wide ranges of time series data. Many approaches have been proposed for automating ML systems. Some of the important methods for building an automated ML systems include automated feature engineering (\cite{katz2016explorekit}), Bayesian Optimization for hyper-parameter tuning (\cite{snoek2012practical}, \cite{bergstra2013making}) and automatic optimization for prespecified sets of of ML models and data pre-processing techniques (\cite{feurer2015efficient}, \cite{kotthoff2017auto}). Moreover, recent research has proposed ways to build fully automated ML systems by incorporating many of the methods mentioned above (\cite{diethe2019continual}).

\section{Algorithm} \label{sec:algorithm}
In this section, we will discuss the step-wise process of building a reliable time series anomaly detection method. It is important to note that although we will discuss univariate time series anomaly detection problems in the modeling section, this approach can be extended to multivariate time series anomaly detection, streaming data anomaly detection, cross-sectional or spatial anomaly detection problems without changing the other proposed system components.

Any machine learning problems can be separated into four major parts: data collection, preparation, modeling (training) and serving (scoring). For the scope of this paper, we will focus on the latter three parts.

\subsection{Data Preparation}
The first step in building a reliable ML system is to develop a process of data cleaning which utilizes techniques that are relevant to the immediate use case. In our case, since we are building an anomaly detection platform for time series data, we need to identify data issues that might leak into our model. Another aspect is batch data profiling which helps us better understand the pattern of the time series in the past. Here are some of the checks and actions that can be applied for preparing a time series before training:
\begin{itemize}
\item \textbf{Imputation and Smoothing:} An important first step before modeling a time series is to perform  missing data imputation, including any missing datetime index or value. This step is important for a time series problem since each data point is sequentially correlated to its neighboring data points. It is also important to smooth the raw time series in some cases in order to obtain valuable information to incorporate into the model.
\item \textbf{Change Point Detection:} A change point is a time point or a collection of time points that shows a change in the distributional properties of the time series. A time series over its history may show few or several change points depending on what the data represents. Change points significantly add to the non-stationarity of the underlying time series, unless there are explainable external events. Hence, it is quite important to identify and take necessary actions in order to avoid any modeling issues.
\item \textbf{Trend Change Detection:} Trend changes are a specific type of change point that might be of particular interest. Changes to past trends can also be important events to be caught for exploratory purposes (such as a shift in market sentiment about a product). These can be considered as events that identify changes happening over a long period of time that have not been individually caught as outliers.
\item \textbf{Stationarization:}  A stationary process is a stochastic process whose statistical properties remains static over time. Typically, time series data shows severe non-stationarity which brings difficulties in modeling. Identification and removal of any non-stationarity is a necessary step before applying any traditional structural or Markovian statistical, or even a machine learning model (unless it is featurized somehow, which a challenging process for building a generic anomaly detector).
\end{itemize}

\subsection{Modeling}
There are two major qualities we need to look for in a time series model built around anomaly detection use cases. The first one is to correctly measure the data uncertainty of the time series; the second being the predictive capacity of the model. It is important to note that measuring and quantifying uncertainty sometimes gets higher priority than predictive capabilities of the model. This is because our purpose is to build an anomaly detection system for time series data with wildly diverse patterns. With this goal, attaining a certain predictive accuracy may not be possible in many situations. Moreover, a lack of training data, external features, and labeled data may make it  difficult to control for any over- or under-fitting of the model if it is solely optimized around predictive capabilities.

There are several standard methods available for modeling time series data. They can be divided into three major groups:
\begin{itemize}
\item \textbf{Feature based Modeling:} These types of models relies on external features to be explicitly added to the model for predictive purposes. Such external features can be specific hours, days or months, holidays or other external events. Examples of such models include RNN, LSTM etc.
\item \textbf{Structural Modeling:} This modeling technique relies on past patterns in features of the time series data to predict future patterns, using information including the trend, periodicity, autocorrelations at different lags, etc. In other words, this class of models uses historical structures or patterns in the time series itself for making predictions. Examples of such models are Auto regressive models, ARMA or ARIMA models etc. Sometimes, external feature information can be combined with a structural model. An example of such a technique is ARIMAX.
\item \textbf{Filter based Modeling:} As discussed earlier, it is sometimes very hard to build a model around predictive accuracy if the data does not show much structural or event based signals. In these cases, it becomes important to quantify the fluctuation patterns to distinguish the anomalous ones from the regular ones. Examples of such techniques include the standard 3-sigma methods, as well as more sophisticated Kalman Filter based modeling.
\end{itemize}

A feature-based model can be greatly optimized around a specific time series under the assumption that the feature set is exhaustive enough to explain the variational patterns. Hence, building an optimal feature-based model requires a great amount of feature engineering, and correspondingly requires a large training dataset to attain a certain degrees of freedom that allows the parametric estimates to achieve a high estimation efficacy. Because of these constraints, our approach focuses on using structural and filter-based modeling over feature-based modeling.

Let us specify the structural model in the following format: \begin{eqnarray}
\label{arimax}
\Phi(B)Y_{t} = \theta(B)X_{t} + \Omega(B)\epsilon_{t}
\end{eqnarray}
Here, $Y_{t}$ is the value of the time series at time $t$; $\Phi(B)$ contains autoregressive terms; $\Omega(B)$ contains the moving average terms; and $\epsilon_{t}$ is a white noise process which overall represents a standard ARMAX model setup. This is to note that we separately quantify the stationarity of the data in the data preparation section to gain more control during the scoring process. Moreover, in order to capture any periodic component present in the data, we represent $X_{t}$ as, \begin{eqnarray}
\label{fourier}
X_{t} &&= \sum_{F=1}^{l}\alpha_{F} e^{2\pi Ft}\\
      &&= \sum_{F=1}^{l}\alpha_{F}\{\cos(2\pi Ft) + i\sin(2\pi Ft)\}
\end{eqnarray}
where $\alpha_{F}$ is the coefficient for the $F^{th}$ frequency term in the Fourier series, and $l$ is the most significant number of Fourier frequencies obtained through the power spectral density \cite{heinzel2002spectrum}.

The exogenous term $X_{t}$ can also include some external periodic features such as holidays or significant events, although significant data with multiple occurrences of all feature events are required. The scope of building such feature based modeling is limited due to the limitations discussed above.
The reason for calling the above model structural is that every piece of structural information (autocorrelation, trend, seasonality, etc.) is extracted from the time series and explicitly used as the modeling components. This works well when the the time series contains good structural signals. Otherwise, we need to introduce an uncertainty based model to quantify the fluctuation patterns. We use the general state space setup for such use cases, \begin{eqnarray}
\label{statespace}
Y_{t} = A_{t}X_{t} + V_{t}
\end{eqnarray}
where $A_{t}$ is the relationship matrix and $V_{t}\sim N(0, R)$ is the measurement noise. Also, \begin{eqnarray}
\label{states}
X_{t+1} = C_{t}X_{t} + W_{t}
\end{eqnarray}
Here $X_{t}$ is the hidden state corresponding to the observed state $Y_{t}$. $C_{t}$ defines the relationship between the consecutive states as well as the off-diagonal elements represents any cross-sectional relationships. $W_{t}\sim N(0,Q)$ represents the process noise for the underlying state propagation.

We use the standard Kalman Filter based technique for quantifying a residual process $\eta_{t+1} = \hat{X}_{t+1\mid t+1} - \hat{X}_{t+1\mid t}$ ($\hat{X}_{t+1\mid t}$ and $\hat{X}_{t+1\mid t+1}$ being the prior and the posterior state estimates respectively) by modeling its variational patterns using a Gaussian process \cite{soule2005combining}.

Sometimes, the time series data shows severe skewness, which is a challenge to capture through a linear modeling setup. This is prevalent in situations where the time series is designed to capture extremities, for example, time series of $p95$ or $p99$ latencies. In those scenarios, defining a multiplicative model (or additive on a logarithmic scale) generates a better fit over the skewness. \begin{eqnarray}
\label{multip}
&&\Phi(B)Y_{t} = \theta(B)X_{t} \cdot \Omega(B)\epsilon_{t}\\
\implies &&\ln{\Phi(B)Y_{t}} = \ln{\theta(B)X_{t}} + \ln{\Omega(B)\epsilon_{t}}\\
\implies &&\Phi^{*}(B)Y_{t}^{*} = \theta^{*}(B)X_{t}^{*} + \Omega^{*}(B)\epsilon_{t}^{*}
\end{eqnarray}

\subsection{Optimization}
The biggest challenge of optimizing the configuration space for an anomaly detection model is the lack of labeled anomalous data. Moreover, temporal dynamicities or pattern shifts invalidate even moderately recent labeled anomalies, and require a continuous flow of labels that makes the process highly unscalable. In general, the traditional time series anomaly detection methods usually rely on forecasting accuracy rather than optimizing the models around their anomaly classification capabilities.

Under the modeling setup discussed in the last few sub-sections, the optimization process involves identifying the actions to perform during the data processing step, data modeling step, and setting any other hyperparameters so that the result is in close proximity to its optimal value. These actions include the decision to truncate the training time series data due to an observed shift in the data property, proportion of missing data to allow in the time series given the amount of signal observed, decision to model the data in log scale, modeling approach to choose between with the corresponding modeling configurations. We can list the configuration ($\psi$) need to be tuned as follows:
\includegraphics[ width=0.48\textwidth]{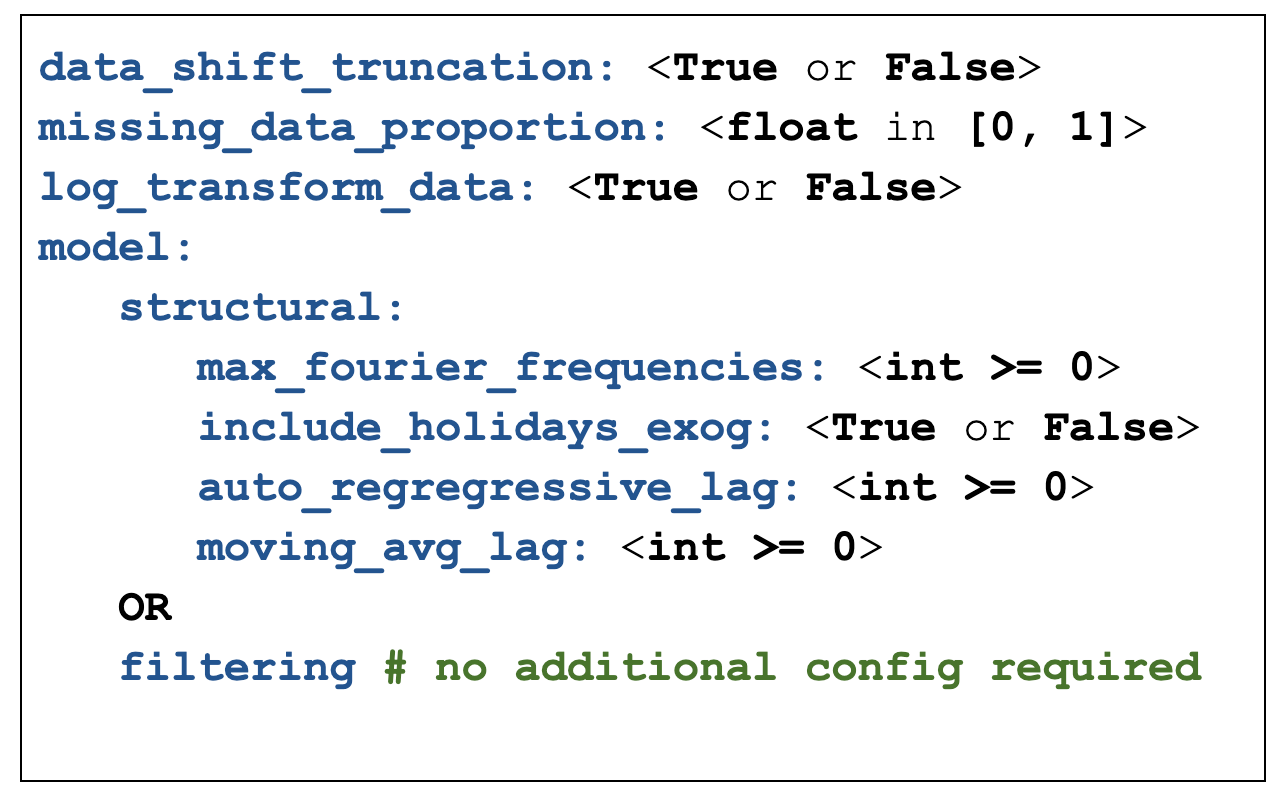}

It is important to note that the configuration space can be extended to a further complex structure by incorporating additional data pre-processing techniques, bringing more complexity to the existing models or even introducing a new modeling method without changing the overall optimization approach.

The above nested configuration space makes it very complex to create a unified cost function for the problem since the structural model is designed to be optimized around forecasting accuracy whereas the filtering model is an uncertainty quantification method around the evolving fluctuation patterns. To build a cost function optimized towards anomaly classification and simultaneously improve the prediction accuracy (whenever possible), we can set up our cost function as the following convex combination of two cost functions: \begin{equation}
\label{cost}
J(\psi) =\\
\begin{cases}
\alpha \textrm{CE} + (1 - \alpha)\cdot \textrm{MAPE}& \small{\text{if\texttt{ method='structural'}}}\\
\textrm{CE}                                         & \small{\text{if\texttt{ method='filtering'}}}
\end{cases}
\end{equation}

Here, $\textrm{CE}$ is the cross-entropy for the observed anomaly probabilities generated from the models and the labeled data with anomalies and non-anomalies. The second part of the cost function for structural models optimizes the predictive accuracy for the selected configuration. This means we optimize the classification accuracy for only the filtering model when the underlying time series carries almost no signal, and perform a weighted optimization on both classification and forecasting accuracy (using mean absolute percentage error) for the structural model when the time series data contains a good amount of signal. Here, $0 \leq \alpha \leq 1$.

In order to generate labeled data for calculating the cross-entropy, we inject randomized synthetic anomalies over the smoothed time series data at different scales. We use tree parzen estimators as a search algorithm for optimizing the configuration space \cite{bergstra2013making}.

We can use the above optimization method to identify the optimal settings required for the mathematical formulation of a model for classifying anomalies. Therefore, the entire workflow including the optimization step, followed by data preparation and modeling steps, all require minimal human intervention.

\begin{figure}[H]
\begin{center}
\includegraphics[ width=0.48\textwidth]{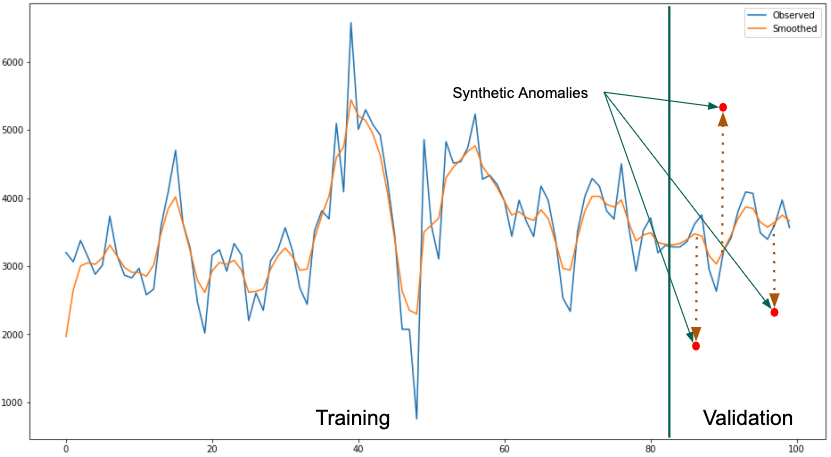}
\caption{Classification optimization through synthetic anomaly injection}
\label{synthetic_anomalies}
\end{center}
\end{figure}

\section{Anomaly Detection System} \label{sec:system}
\begin{figure*}
\begin{center}
\includegraphics[ width=\textwidth]{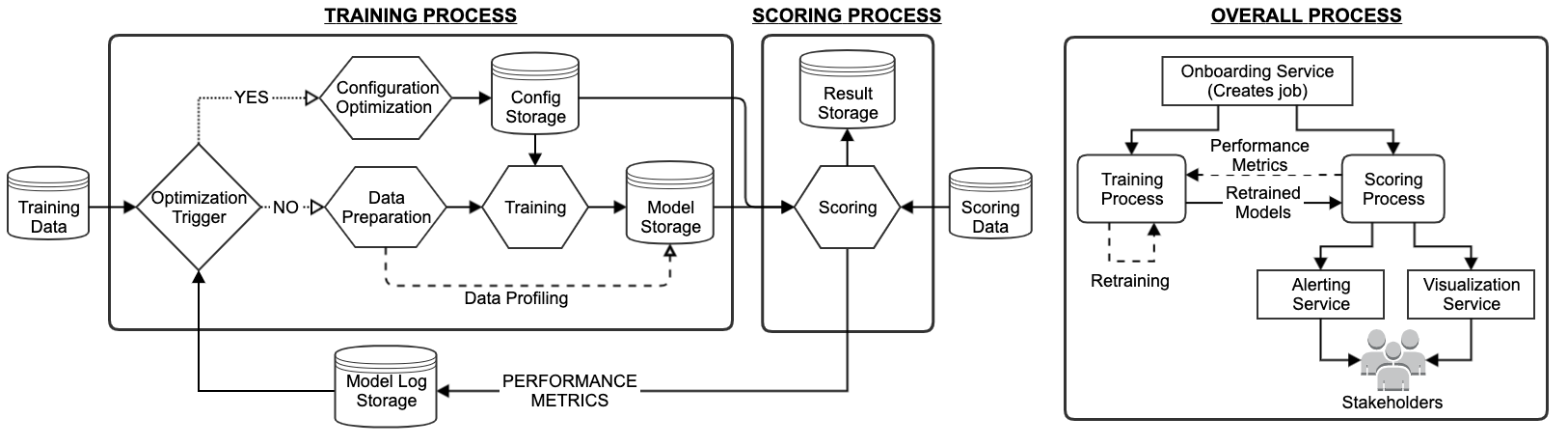}
\caption{Architecture for the proposed automated and self-aware anomaly detection system}
\label{full_workflow}
\end{center}
\end{figure*}

Our target is to build an automated and self-aware anomaly detection system which can run at scale. The section above has mostly covered considerations around minimizing the configuration requirement for the system. Further automation is still required to enable the system to run autonomously, and to keep track of its own performance in order to fine-tune itself over time. In this section, we will discuss how we can combine schedule and trigger based processes in order to create a fully hands-off anomaly detection system.

\subsection{Data Acquisition}
The very first step for the system is the ability to acquire the data. Data acquisition can be achieved in two ways, data pull and data push approach. In data pull, system should be able to connect and query the database where data resides. In data push, system should provide an end-point where process can pass data to it.

In both the approach, system should be able support monitoring of multiple anomaly detection jobs. For this, system should maintain job and metrics level information, and assign distinct identifiers. Distinct identifiers becomes the stepping stone for linking all the data system generate for each of those monitoring job and metrics in it.

\subsection{Training}
The training approach we discuss is a general setup that applies to any anomaly detection system, but there are some specifics about scheduling the training process that are special to the time series use case. A key difference between a model over cross-sectional data and a model over sequential data is that sequential models require very systematic training processes due to inherent non-stationarity. We can consider a scheduling engine that triggers the training process on a regular interval. A fully automated training process can be designed as shown in figure \ref{full_workflow}, where we sequence the sub-processes of configuration tuning, followed by data preparation and cleaning, and finally the actual training of the time series.

We consider a model storage that can store time series models after training, including all metadata such as a unique model identifiers, information related to when the model is published, and any expiry policies. Defining model expiration is an important step for time series data due to non-stationarity. Scheduling the training process varies significantly based on the data type or the data frequency. For example, high frequency data tend to show better stationarity compared to low frequency data given the same amount of innovation (e.g. a time series of length 30 with innovations per minute tend to show better stationary properties compared to a similar length time series with innovations per day).

\subsection{Scoring}
In the scoring process, we score new temporal innovations using existing trained models that are stored. The scoring process in general runs on a different schedule more frequently than the training process. Usually scoring should run at the frequency that the metrics getting monitored are generated at. For example, to monitor page views of a website that is aggregated every hour, we can train a model once in two days and can use that model to score every hour for next $48$ hours of data. The scoring process obtains the right model from the model storage using the unique identifier, classifies the new innovation to be stable or anomalous using a probability value and a pre-specified threshold, and outputs its decision to the result database (shown in figure \ref{full_workflow}).

The last critical component to anomaly detection is alerting stakeholders for metrics they care about which are identified as anomalous by the scoring process. This can be achieved by creating an Alerting Service. Alerting to stakeholders can be achieved via various alerting channels such as email, chat based apps, ticketing systems, and on-call systems.

\subsection{Performance Evaluation}
Evaluating the performance of a machine learning system becomes challenging under an unsupervised setup. For anomaly detection problems, traditional ways of evaluating the system with Receiver Operating Characteristic (ROC) is not possible due to the absence of labeled data. In this paper, we use the ranked probability scores to evaluate an anomaly classification model (\cite{goix2016evaluate}, \cite{goix2015anomaly}).

\begin{figure}[H]
\begin{center}
\includegraphics[ width=0.49\textwidth]{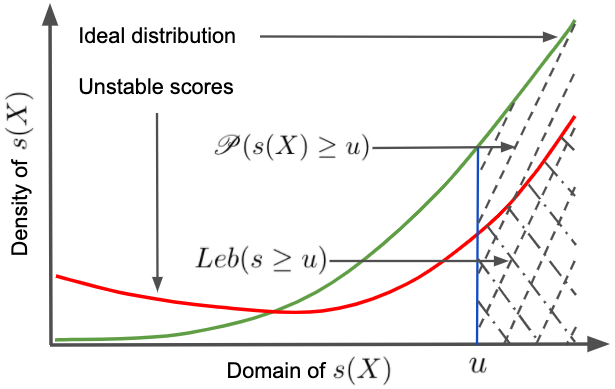}
\caption{Comparison between ideal and unstable states of an anomaly detector}
\label{evaluation}
\end{center}
\end{figure}

Let us consider $\tau_{1}, \tau_{2}, ..., \tau_{k}$ be the set of $k$ time series being monitored within a task. We define a scoring function $s_{i}: \mathscr{R}^{d} \longrightarrow  \mathscr{R}$ where $d = 1$ for our univariate time series scenario, $i = 1,2,...k$ and $s_{i}$ is an instance from the set of all scoring functions $\mathscr{S}$. We can consider $s_{i}$'s to represent the degree of abnormality for a point $x^{(i)}\in\Omega^{(i)}\subseteq \mathscr{R}^{d}$. If we assume $f$ be the true representation for the data density on a unified transformed domain of the time series $\tau_{i}$ $\forall i = 1, 2, . . ., k$ represented by $x^{*(i)}\in\Omega^{*}\subseteq \mathscr{R}^{d^{*}}$, then for the ideal scoring function $s$ defined over the same transformed domain $\Omega^{*}$, which is an optimal representation of stability and abnormality, we can create a criterion based on $\mathscr{C}^{\Phi}(s) = \left\Vert\Phi(s) - \Phi(f)\right\Vert$ where $\Phi: \mathscr{R}\longrightarrow \mathscr{R}_{+}$ is either the Mass Volume $(\mathrm{MV}_{s})$ curve or the Excess Mass $(\mathrm{EM}_{s})$ curve of $s$ defined as: \begin{eqnarray}
&&\mathrm{MV}_{s}(\alpha) = \inf_{u\geq 0} Leb(s \geq u)\hspace{0.1cm}s.t.\hspace{0.1cm} \mathscr{P}(s(X)\geq u)\geq \alpha \hspace{0.5cm}\\
&&\mathrm{EM}_{s}(t) = \sup_{u\geq 0} \mathscr{P}(s(X)\geq u) - t\cdot Leb(s\geq u)
\end{eqnarray}
where $s(\cdot)$ is a scoring function integrable with respect to any Lebesgue measure $Leb(\cdot)$. The idea behind the above formulation is the assumption that anomalies occur in the low probability region in the data domain and hence the scoring function should have a small value. On the other hand, stable data comes from a high probability region in the data domain and the scoring function should take higher values. A better illustration of the idea can be observed in figure \ref{evaluation}. We can identify an unstable or under-performing configuration setup if the scoring distribution starts tilting to the right from the ideal situation (this requires keeping the logs of the scores when it satisfies a set of pre-specified conditions of stability). The optimal curves are $\mathrm{MV}^{*} = \mathrm{MV}_{f}$ and $\mathrm{EM}^{*} = \mathrm{EM}_{f}$ and it can be shown that \begin{eqnarray}
&&\mathrm{MV}^{*}(\alpha) \leq \mathrm{MV}_{s}(\alpha)\\
&&\mathrm{EM}^{*}(t) \geq \mathrm{EM}_{s}(t)
\end{eqnarray}

Therefore, for a stable model and configuration setup, the corresponding scores are expected to produce relatively low values for the $\mathrm{MV}$ function and relatively high values for the $\mathrm{EM}$ function. The final performance criteria is computed based on the averages of the $\mathrm{MV}$ and the $\mathrm{EM}$ function under certain pre-specified domains for $\alpha$ and $t$ \cite{goix2016evaluate}. In practice, we consider, \begin{eqnarray}
s(x) = 1-\mathrm{AnomalyProbability}(x) 
\end{eqnarray}
across all the time series $\tau_{1}, \tau_{2}, ..., \tau_{k}$ to obtain an ideal estimated density for the underlying anomaly level distribution for a task under certain data stability conditions and compare the individual $s_{i}(x^{(i)})$ with the ideal anomaly level density to evaluate the performance of the $i^{th}$ time series model and configuration.

\subsection{Logging Performance Metrics and Self-Awareness}
A very important step for the system to self-monitor is to keep a log of its performance. Time series data evolves in different ways over time including shifts, variational patterns, evolving temporal correlational structure, and changing periodicities. Any new pattern that has never been observed in the data is treated as anomalous, but if the change is  sustained, the model and/or the underlying configuration should evolve with it. An important checkpoint to register any new change in the innovation is the scoring information. Any sub-optimal performance of the model due to older configuration or new patterns should be logged and these information can be used to re-trigger the tuning process.

We can consider some basic performance indicators, such as time series that are consecutively or frequently anomalous, or time series that are having training failures due to sub-optimal or invalid configuration. It is important to understand that such indicators are collected at the cost of anomaly alerts or failures which may not be an optimal experience for the end user of the anomaly detection service. Taking corrective measures after sending too many alerts can create alert fatigue. Failing to alert for a critical issue due to model failures can result in serious consequences. Therefore, we need an indicator of performance deterioration that can trigger an action before the service generates any bad outputs. We can consider collecting the MV's and EM's across all the time series as discussed in the previous subsection, and can identify a time series to require a tuning based on a threshold criterion for both MV and EM. Moreover, we can collect indicators such as Coefficient of Variation or model freshness (whether a time series is being scored using an almost expiring model) to trigger  configuration updates.

At every training cycle of the data, we can consider three mutually exclusive sets of time series indicating with green ($G$: set of healthy time series model configuration), yellow ($Y$: set of time series model configurations under scrutiny) and red ($R$: set of time series models that need configuration tuning) and run the optimization trigger for the set $R$ (as shown in figure \ref{full_workflow}). \begin{eqnarray}
G = \{\tau_{i}^{G}: i=1,2,...,k_{1}\}\hspace{0.2cm}\\
Y = \{\tau_{i}^{Y}: i=1,2,...,k_{2}\}\hspace{0.2cm}\\
R = \{\tau_{i}^{R}: i=1,2,...,k_{3}\}\hspace{0.2cm}\\
G\cup Y\cup R = \{\tau_{i}: i=1,2,...,k\}
\end{eqnarray}

Although the idea of performance evaluation and self-awareness has been proposed in this paper around anomaly detection models in general, a similar idea can be extended to any ML system where performance metrics can be logged and retrieved during the workflow. For example, one can log the performance metrics around predictive accuracy such as MAPE or RMSE to build a self-aware ML system for forecasting.

\section{Stable and Scalable system} \label{sec:scaling}
For an organization to achieve broad adoption of an anomaly detection system, we recommend implementing several engineering practices to enable proper scalability and stability of the system. The below practices provide support for many concurrent jobs with efficient use of compute resources for training and scoring processes and support the high availability and stability of the system itself.

\subsection{Distributed Processing}
Each anomaly detection job can consist of a collection of metrics to track since stakeholders typically consider a group of metrics about a dataset or service together. For example, the number of visits to a website may be tracked in a metric collection, broken down by browser. When represented as a collection, many metrics can be trained or scored efficiently in parallel using a distributed compute framework, such as Apache Spark.

\subsection{Auto Scaling}
The load on the anomaly detection system can vary by days of the week and/or hour of the day. Having a fixed environment of processing resources has two drawbacks. First, it cannot meet growing resource demands as more metrics are added, requiring manual infrastructure changes when the environment is outgrown. Additionally, large, long-running environments waste resources during periods of light system load. We recommend setting up auto scaling policies which can increase or decrease the system resources based on the demand on the system. Auto scaling can be based on various system metrics, such as resources utilized and jobs pending.

\subsection{Monitoring}
To be able to rely on a data quality system to catch issues in a timely and reliable manner, the health of the system itself must be monitored in a robust way. We suggest monitoring a number of system metrics, including system availability, failed connections, resource utilization, unhealthy nodes, number of jobs failed, jobs killed, jobs pending and job run time. Deterministic thresholds can be configured on top of these metrics and connected to an organization's internal tools for alerting, ticketing, service level agreements, and on-call support.

\section{Benchmarks} \label{sec:benchmarks}

\subsection{Anomaly Classification}

\begin{table*}[!b]
\caption{AUC comparison with NAB benchmark datasets}
\label{auc_comparison}
\noindent\begin{center}
\scalebox{1}{\begin{tabular}{|p{5cm}|p{1cm}|p{1cm}|p{1cm}|p{1cm}|p{1cm}|p{1cm}|p{1.4cm}|p{1.4cm}|}
\cline{2-9}
\multicolumn{1}{c|}{}& \multicolumn{8}{c|}{Methods} \\
\cline{2-9}
\multicolumn{1}{c|}{}& \multicolumn{2}{c|}{Prophet} & \multicolumn{2}{c|}{luminol} & \multicolumn{2}{c|}{ADTK} & \multicolumn{2}{c|}{AutoAD}\\
\hline
Datasets & Daily & Hourly & Daily & Hourly & Daily & Hourly & Daily & Hourly \\
\hline
Twitter\_volume\_CRM    & 0.67287  & 0.64670 & 0.51223  & 0.62413 & 0.62131 & 0.58854 & \textbf{0.75267} (3) & \textbf{0.87414} (4) \\
\hline
Twitter\_volume\_FB       & 0.43889 & 0.65513 & 0.46233 & 0.32812 & 0.68889 & 0.48214 & \textbf{0.70909} (1) & \textbf{0.76227} (3) \\
\hline
Twitter\_volume\_GOOG    & 0.66433  & 0.60139 &  0.57244 & 0.61250  & 0.68027 &0.51250 & \textbf{0.72889} (3) &\textbf{0.76667} (3) \\
\hline
nyc\_taxi   & 0.60606 & \textbf{0.87413} &  \textbf{0.68687} & 0.50233 & 0.64141 & 0.73958 & 0.65151 (3) & 0.72483 (5) \\
\hline
machine\_temparature\_system\_failure    & 0.93333 & 0.98839 & 0.79333 & 0.88710 & 0.58333 & 0.52984 & \textbf{0.96787} (2) & \textbf{0.99623} (3)\\
\hline
cpu\_utilization\_asg\_misconfiguration    &  0.63333 & 0.51250 & 0.32917 & \textbf{0.74861} & \textbf{0.74167} & 0.50833 & 0.72667 (3) & 0.44167 (3) \\
\hline
\end{tabular}}
\end{center}
Number of configuration retunings for AutoAD is shown in parentheses in the corresponding columns
\end{table*}

In this section, we consider comparing the anomaly classification performance of the proposed automated anomaly detection system (AutoAD) against different existing time series anomaly detection solutions. We consider NAB (The Numenta Anomaly Benchmark) datasets for comparison purpose \cite{lavin2015evaluating}. Although the proposed self-aware and automated anomaly detection architecture can be extended to various spatial, spatio-temporal and even for streaming use cases, we consider only real time anomaly detection use cases for this comparison purpose (hourly and daily monitoring). We perform aggregations over some selected NAB datasets with decent amount of history (on an aggregated level) since the original data is observed at a higher frequency (every 5 minutes or 10 minutes)

Since we have proposed a real time anomaly detection system, we consider the system with the following three solutions which are widely used and are built towards time series anomaly detection and forecasting:
\begin{itemize}
\item \textbf{Prophet\footnote[1]{Prophet project link: \url{https://facebook.github.io/prophet/}}:} An automated time series forecasting procedure \cite{taylor2018forecasting}.
\item \textbf{luminol\footnote[2]{luminol project link: \url{https://github.com/linkedin/luminol}}:} A light-weight time series anomaly detection method.
\item \textbf{ADTK\footnote[3]{ADTK project link: \url{https://github.com/arundo/adtk}}:} An unsupervised/rule based time series anomaly detection system.
\end{itemize}

For our comparison purpose, we consider hourly and daily aggregations of the following benchmark datasets available in NAB:
\begin{itemize}
\item Twitter\_volume\_CRM
\item Twitter\_volume\_FB
\item Twitter\_volume\_GOOG
\item nyc\_taxi
\item machine\_temparature\_system\_failure
\item cpu\_utilization\_asg\_misconfiguration
\end{itemize}

\begin{figure}[H]
\begin{center}
\includegraphics[ width=0.48\textwidth]{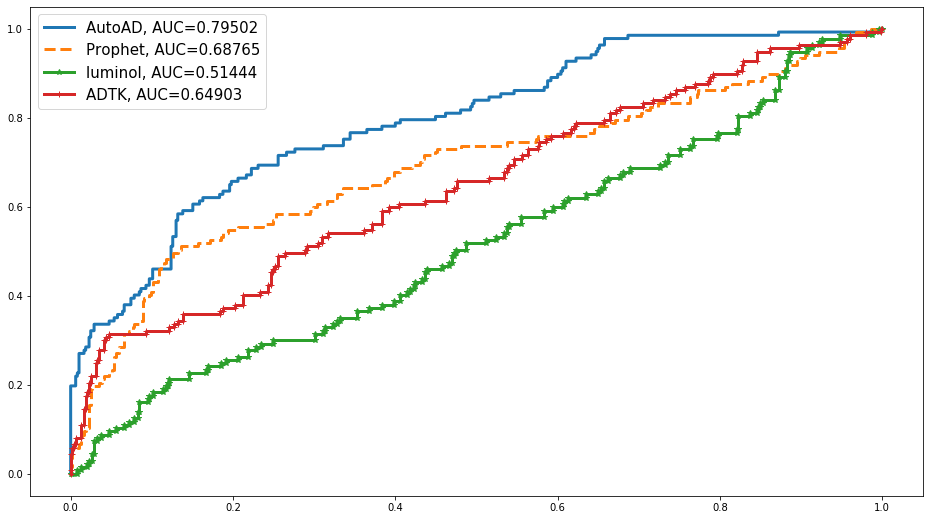}
\caption{ROC curve for NAB daily and hourly aggregated benchmark datasets}
\label{daily_roc}
\end{center}
\end{figure}

For each of these datasets, we sequentially train (less frequently) and score (for every timestamp) and let AutoAD to auto-trigger configuration retuning whenever necessary. For the competing methods, we either perform training at a same schedule (for Prophet) and use the trained model to score nearest future timestamps or perform batch anomaly detection (for luminol and ADTK).

In the Area under an ROC curve (AUC) comparison from table \ref{auc_comparison}, we can see that the proposed AutoAD method outperforms the other methods in many scenarios. Moreover, AutoAD tends to generate more consistent performance across all the datasets compared to the other methods. We also generate a combined ROC curve by combining all the true labels and the predictions from all the datasets in figure \ref{daily_roc}. This ROC shows that the AutoAD consistently outperforms the other methods mostly throughout the x axis of the curve. This means the AutoAD performs well in terms of minimizing false positive rate as well as maximizing the true positive rate.

This is to note that the anomaly classification is achieved along with the level of automation in the system. In other words, the automation and the maintenance strategy built in the system actually helps the modeling workflow to properly digest the wide temporal variations and patterns across different data over time. Specifically, the configuration optimization strategy allows the training process to search over the entire parameter space to build a custom model for a given data whenever needed.

\subsection{Forecasting Performance}

A key component of a time series anomaly detection tool is its ability to forecast. If the underlying time series contains good enough signal, the performance around anomaly classification becomes highly correlated with forecasting performance. Equation \ref{cost} shows that the optimization method is actually built around a weighted combination of classification and forecasting capabilities of the model. Therefore, the proposed automated anomaly detection method can also be used as a forecasting tool which is another important ML tool for a stable machine learning system. For example, a daily anomaly detection model build to track cluster resources for possible anomalies can also be used for predicting future use of resources for budgeting purpose. The value of $\alpha$ from equation \ref{cost} can be tweaked based on the modeling requirements. Someone can set an alpha greater than $0.5$ if the optimization should be done around anomaly detection and can be set less than $0.5$ if it needs to be optimized for forecasting. Note that setting it near to either $0$ or $1$ may induce underperformance due to overfitting or underfitting with respect to the forecasting capabilities.

\begin{table*}[t]
\caption{Forecasting performance comparison with NAB benchmark datasets with Median Absolute Percentage Error (MDAPE) and Root Mean Squared Error (RMSE).}
\label{forecast_comparison}
\noindent\begin{center}
\scalebox{0.85}{\begin{tabular}{|p{4.2cm}|p{0.93cm}|p{0.93cm}|p{0.93cm}|p{0.93cm}|p{0.93cm}|p{0.93cm}|p{0.93cm}|p{0.93cm}|p{0.93cm}|p{0.92cm}|p{0.93cm}|p{0.93cm}|}
\cline{2-13}
\multicolumn{1}{c|}{}& \multicolumn{12}{c|}{Methods} \\
\cline{2-13}
\multicolumn{1}{c|}{}& \multicolumn{4}{c|}{Prophet} & \multicolumn{4}{c|}{auto\_arima} & \multicolumn{4}{c|}{AutoAD}\\
\cline{2-13}
\multicolumn{1}{c|}{}& \multicolumn{2}{c|}{Daily} & \multicolumn{2}{c|}{Hourly} & \multicolumn{2}{c|}{Daily} & \multicolumn{2}{c|}{Hourly} & \multicolumn{2}{c|}{Daily} & \multicolumn{2}{c|}{Hourly}\\
\hline
Datasets & MDAPE & RMSE & MDAPE & RMSE & MDAPE & RMSE & MDAPE & RMSE & MDAPE & RMSE & MDAPE & RMSE \\
\hline
Twitter\_volume\_CRM    & 57.659\% & 550.419 & 51.689\% & 45.997 & 40.835\% & 450.688 & 54.524\% & \textbf{27.677} & \textbf{28.572\%} & \textbf{396.090} & \textbf{54.340}\% & 30.922 \\
\hline
Twitter\_volume\_FB     & 14.287\% & 1098.187 & 43.691\% & 139.053 & 16.951\% & \textbf{1048.337} & 40.719\% & 122.255 & \textbf{11.137\%} & 1119.641 & \textbf{35.468\%} & \textbf{96.602} \\
\hline
Twitter\_volume\_GOOG   & 19.830\% & \textbf{1380.836} & 52.916\% & 179.444 & \textbf{16.607\%} & 1619.820 & 40.684\% & 123.197 & 24.004\% & 1619.16 & \textbf{27.877\%} & \textbf{96.107} \\
\hline
nyc\_taxi  & \textbf{2.135\%} & \textbf{37595.87} & 27.015\% & 14614.37 & 5.061\% & 50913.99 & 27.403\% & 12024.55 & 5.664\% & 79176.01 & \textbf{15.728\%} & \textbf{8499.727} \\
\hline
machine\_temparature\_system\_failure   & 12.808\% & 3690.954 & 8.893\% & 154.710 & 8.049\% & \textbf{2079.783} & 4.362\% & 98.074 & \textbf{4.855\%} & 2134.686 & \textbf{3.735\%} & \textbf{96.794} \\
\hline
cpu\_utilization\_asg\_misconfiguration  & \textbf{4.717\%} & \textbf{574.292} & 6.562\% & 55.538 & 6.672\% & 817.155 & \textbf{5.184\%} & \textbf{28.292} & 5.555\% & 675.602 & 7.112\% & 37.017 \\
\hline
\end{tabular}}
\end{center}
Number of configuration retunings for AutoAD is same as shown in Table \ref{auc_comparison}
\end{table*}

\begin{table*}[!b]
\caption{Average run-time per training (for a single time series) comparison using simulated data of different lengths (The competing methods have only one run-time reported for each time series length scenario as there is no concept of Optimization trigger)}
\label{runtime_comparison}
\noindent\begin{center}
\scalebox{0.9}{\begin{tabular}{|p{1cm}|p{1.1cm}|p{1.1cm}|p{1.1cm}|p{1.1cm}|p{1.1cm}|p{1.1cm}|p{1.1cm}|p{1.1cm}|p{1.1cm}|p{1.1cm}|p{1.1cm}|p{1.1cm}|}
\cline{2-13}
\multicolumn{1}{c|}{}& \multicolumn{4}{c|}{Time series of length 1k} & \multicolumn{4}{c|}{Time series of length 2k} & \multicolumn{4}{c|}{Time series of length 3k}\\
\hline
Methods & 0 Opt & 1 Opt & 2 Opt & 3 Opt & 0 Opt & 1 Opt & 2 Opt & 3 Opt & 0 Opt & 1 Opt & 2 Opt & 3 Opt \\
\hline
AutoAD    & 2.759 sec & 3.619 sec & 4.962 sec & 7.890 sec & 3.881 sec & 4.694 sec & 6.122 sec & 8.118 sec & 4.252 sec & 5.620 sec & 7.587 sec & 8.477 sec \\
\hline
Prophet   & 2.015 sec & NA & NA & NA & 2.189 sec & NA & NA & NA & 3.251 sec & NA & NA & NA \\
\hline
luminol   & 0.030 sec & NA & NA & NA & 0.055 sec & NA & NA & NA & 0.062 sec & NA & NA & NA \\
\hline
ADTK      & 0.038 sec & NA & NA & NA & 0.042 sec & NA & NA & NA & 0.053 sec & NA & NA & NA  \\
\hline
\end{tabular}}
\end{center}
Resource used - Processor: 2.8 GHz Quad-Core Intel Core i7, Memory: 16Gb 2133 MHz LPDDR3;

Opt - Optimization triggers, NA - Not Applicable
\end{table*}

We consider the NAB benchmark datasets for forecasting the proposed method against other competing solutions. Since we have the anomaly labels present in each of the datasets, we consider predicting the region of the time series that are non-anomalous. For the comparison purpose, we use Prophet and pyramid auto\_arima\footnote[1]{auto\_arima project link: \url{http://alkaline-ml.com/pmdarima/0.9.0/modules/generated/pyramid.arima.auto_arima.html}} for competing forecasting methods. We have used daily and hourly scenarios similar to the anomaly detection benchmarking and we compared the predictions using 7 days ahead predictions for daily aggregations and 24 hours ahead predictions for hourly aggregations. We use two different prediction evaluation metrics to compare the performance. The first one is the Median Absolute Percentage Error (or MDAPE) and the second metric is the Root Mean Squared Error (or RMSE). The reason to choose MDAPE over MAPE (Mean Absolute Percentage Error) is to understand a robust representation of the performance in case of an outlying prediction error due to observation or predictions issues. this is also to note that the performance evaluation has been generated using $\alpha = 0.5$ in equation \ref{cost}. Although, the value of alpha can be tweaked based on the modeling requirements.

From table \ref{forecast_comparison}, we can clearly observe that the $AutoAD$ method outperforms the competing forecasting solutions in multiple datasets. In particular, we observe that the performance is significantly better in hourly aggregations compared to daily aggregations. One reason for this effect may be due to the data being much noisier when it is aggregated hourly compared with daily. $AutoAD$ performed significantly better than the other methods for those cases, and may be a promising tool to forecast noisy time series- a very common scenario in Big Data systems.

\subsection{Compute Time}

The modeling process of the proposed anomaly detection system in this paper consists of an optional (or trigger based) configuration optimization step, a data preparation and profiling step and the main training step. The optimization step is a process that should occur infrequently, as it is the most expensive process consisting of Gaussian Process optimization. Although the configuration space under consideration is low dimensional, it is important to understand and compare the compute time with other available solutions to get an idea of the compute performance of the proposed fully automated anomaly detection method with respect to the other solutions which may require manual human interventions for maintenance.

We compare the end to end run-time comparison of AutoAD with the training and scoring process for Prophet, luminol and ADTK. For comparison, we create simulated set of time series data of different lengths and artificially changing the structural properties of the data in order to auto trigger the configuration optimization step for AutoAD. We compare the average run-time per iteration for the competing methods with different scenarios. The end to end  system for AutoAD considers a trigger based optimization step (with a pre-specified convergence criterion), the data preparation and profiling step and the training step. The training has been done for every increment of $k$ new data points with a set of pre-specified values of $k$. For the competing methods, we are only considering the training without updating any configurations manually if required (e.g. truncating the data, taking a transformation of the data to check if we obtain a better fir for the model etc.). In terms of changing the data properties, we consider shifting the process mean and variance, adding or removing trends, changing the periodicity cycles etc.

Although the comparison has been performed on a single machine, this still conveys valuable information about the compute performance at an atomic level with respect to the other methods since, if necessary, the proposed system is easily be scaled up at a cluster level as discussed in section \ref{sec:scaling}.

Table \ref{runtime_comparison} shows the performance comparison between different solutions. We can see that for different lengths of the time series, the compute time for the complete end to end process for AutoAD shows good enough performance and the amount of increase in the compute time is within a significantly small neighborhood from the competing method. Therefore, the fully automated anomaly detection system shows promising performance on classifying anomalies (as been observed before) along with a negligible compute cost compared to the requirement of manual human interventions which has many uncertainty related to time and resources.

\section{Discussion}
This paper proposed a complete methodology for building a time series anomaly detection system that requires very minimal configuration and maintenance. The data preparation and profiling step performs all necessary operations on the input time series data, including imputations, transformations and all other adjustments to prepare the data for the modeling phase. The modeling phase trains the time series data using different modeling approaches ranging from structure based or filter based. Traditionally, the data preparation / profiling model training steps require rigorous configurations that demand extensive domain or ML expertise that may vary significantly between different datasets. Therefore, this paper proposes an optional step for optimizing the configuration for each specific dataset under consideration.

In order to reduce effort due to model maintenance, we propose a number of performance evaluation and logging techniques. The system continuously evaluates its own performance using metrics such as EM, MV, failures, etc. and logs all performance information at the scoring phase. The system then utilizes these logs during the training phase to decide whether the data under consideration requires re-tuning the configuration prior to training the model.

We finally compare the proposed method with some existing time series anomaly detection solutions and show that the proposed system outperforms the existing methods in multiple benchmark datasets. We find that the proposed approach also leads in terms of aggregated anomaly classification performance and forecasting abilities without requiring significant compute time compared to manual interventions.

While the system proposed in the paper is designed to build an automated and self-aware anomaly detection system, this idea can also be extended for building other types of machine learning systems. For example, the modeling approach can be replaced for a self-aware system that can track anomalies in streaming data or the modeling and evaluation system can be restructured to support a recommender system.

\bibliographystyle{IEEEtran}
\bibliography{manuscript.bib}
\end{document}